\documentclass[10pt,twocolumn,letterpaper]{article}

\usepackage[pagenumbers]{cvpr} 

\usepackage{booktabs}
\usepackage{multirow}
\usepackage{makecell}
\usepackage{xcolor}
\usepackage{epsfig}
\usepackage{tikz}
\usetikzlibrary{arrows.meta, positioning, calc}
\newcommand{\gain}[1]{\textcolor{red}{\scriptsize\, (#1)}}
\newcommand{\best}[1]{\textbf{#1}}

\usepackage{url}

\definecolor{cvprblue}{rgb}{0.21,0.49,0.74}
\usepackage[pagebackref,breaklinks,colorlinks,allcolors=cvprblue]{hyperref}

\def\paperID{} 
\def\confName{CVPR}
\def\confYear{2026}

\title{DetRefiner: Model-Agnostic Detection Refinement with Feature Fusion Transformer}

\author{
Soichiro Okazaki$^{1}$ \quad
Tatsuya Sasaki$^{1}$ \quad
Hiroki Ohashi$^{1}$ \quad \\
$^{1}$Hitachi, Ltd. Research and Development Group, Japan\\
{\tt\small \{soichiro.okazaki.xs, tatsuya.sasaki.gb, hiroki.ohashi.uo\}@hitachi.com}
}

\begin{document}
\maketitle
\begin{abstract}
Open-vocabulary object detection (OVOD) aims to detect both seen and unseen categories, yet existing methods often struggle to generalize to novel objects due to limited integration of global and local contextual cues. We propose \textbf{DetRefiner}, a simple yet effective plug-and-play framework that learns to fuse global and local features to refine open-vocabulary detection. DetRefiner processes global image features and patch-level image features from foundational models (e.g., DINOv3) through a lightweight Transformer encoder. The encoder produces a class vector capturing image-level attributes and patch vectors representing local region attributes, from which attribute reliability is inferred to recalibrate the base model’s confidence. Notably, DetRefiner is trained independently of the base OVOD model, requiring neither access to its internal features nor retraining.
At inference, it operates solely on the base detector’s predictions, producing auxiliary calibration scores that are merged with the base detector’s scores to yield the final refined confidence. Despite this simplicity, DetRefiner consistently enhances multiple OVOD models across COCO, LVIS, ODinW13, and Pascal VOC, achieving gains of up to +10.1 AP on novel categories. These results highlight that learning to fuse global and local representations offers a powerful and general mechanism for advancing open-world object detection. Our codes and models are available at \url{https://github.com/hitachi-rd-cv/detrefiner}.
\end{abstract}    
\begin{figure}[t]
  \centering
  \includegraphics[width=0.8\linewidth]{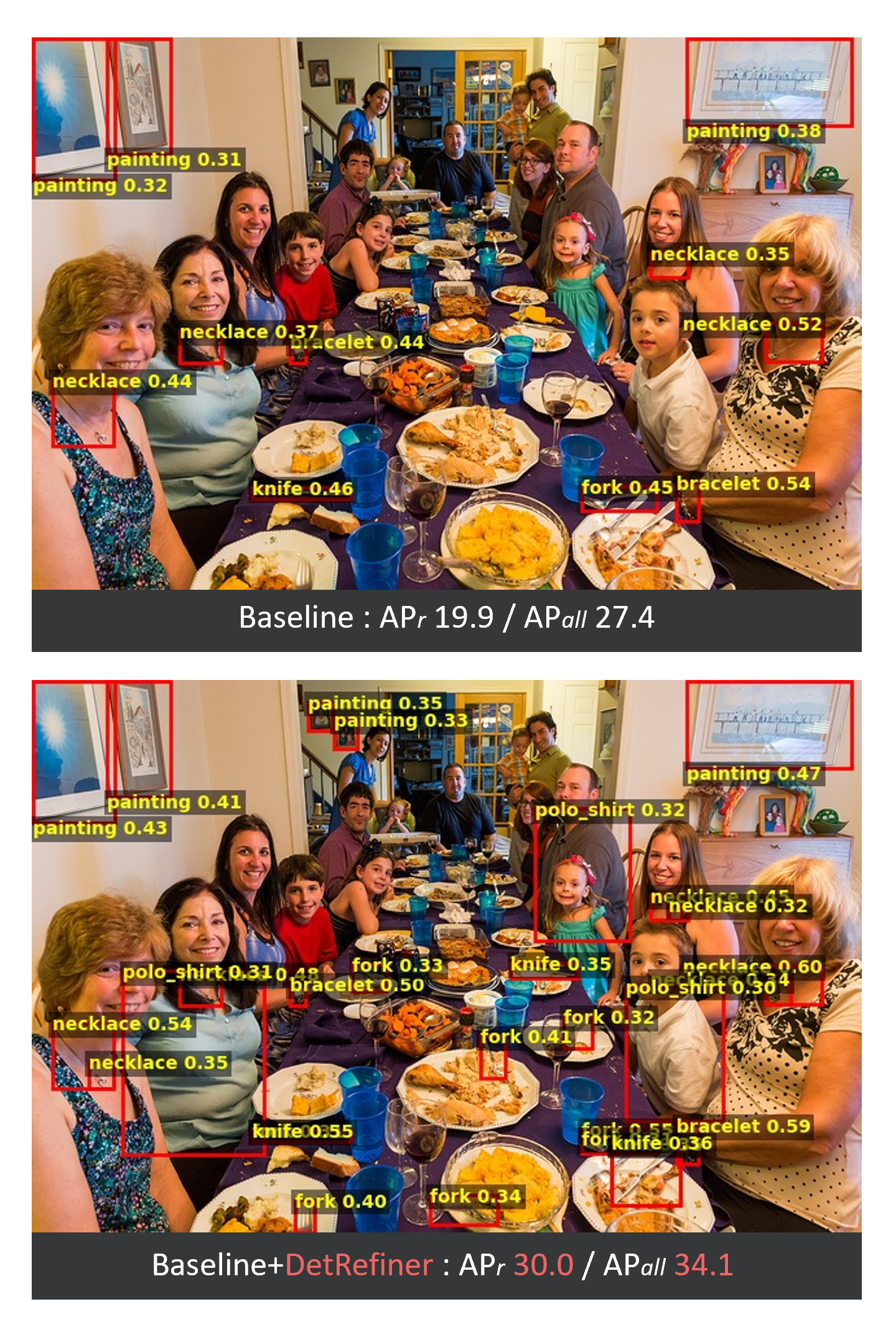}
  \caption{
    Qualitative comparison on \textbf{LVIS} between the baseline detector 
    (Grounding DINO-T~\cite{liu2024grounding}) and the proposed \textbf{DetRefiner}. 
    DetRefiner significantly improves detection performance
    (AP$_r$: 19.9 $\rightarrow$ 30.0 / AP$_{all}$: 27.4 $\rightarrow$ 34.1), successfully detecting more objects and producing better calibrated scores (e.g., \textit{fork}, \textit{knife}, \textit{painting}, \textit{polo\_shirt}). For visualization, a box score threshold of 0.3 and an IoU threshold of 0.3 are applied for class-wise NMS.
  }
  \vspace{-0.6em}
  \label{fig:lvis_detrefiner}
\end{figure}

\section{Introduction}
\label{sec:intro}
Object detection aims to localize and categorize visual instances by predicting bounding boxes and class labels.
Conventional detectors~\cite{ren2016faster,bochkovskiy2020yolov4} and more recent Transformer-based models~\cite{carion2020end,zhangdino} have achieved remarkable progress by learning powerful visual representations and global context modeling.
However, their label spaces are typically \emph{closed}, constrained to a fixed set of categories defined during training.
This closed-world assumption severely limits scalability in practical applications, where new concepts continually emerge, long-tail categories are common, and unseen objects frequently appear in the wild~\cite{gupta2019lvis,wang2023v3det}.

Open-vocabulary object detection (OVOD) addresses this limitation by coupling region localization with text-based classification, typically leveraging large-scale vision--language pretraining (VLP) models~\cite{radford2021learning,jia2021scaling}.
These models learn joint image--text representations from web-scale data and enable category generalization beyond the training label set.
Building on this foundation, methods such as ViLD~\cite{guopen} and RegionCLIP~\cite{zhong2022regionclip} extend pretrained embeddings to the detection setting, aligning region proposals with textual concepts to recognize unseen categories.
More recent approaches including GLIP~\cite{li2022grounded} and Grounding DINO~\cite{liu2024grounding} jointly optimize grounding and detection for stronger open-vocabulary generalization, allowing categories to be specified at inference time as natural-language prompts.

Despite this flexibility, OVOD models still struggle with \emph{semantic alignment} and \emph{score calibration}~\cite{wu2023aligning,shi2023edadet,zheng2024training,wang2024open}. The image--text representations learned via pretraining may misalign with a detector's region embeddings, making it difficult to reliably match visual features to textual descriptions~\cite{wu2023aligning,shi2023edadet}. Furthermore, detectors frequently encounter ambiguous or fine-grained categories, where visually similar objects correspond to semantically overlapping prompts. In such cases, confidence scores often become unstable and poorly calibrated: high scores may be assigned to visually or semantically incorrect detections, while correct predictions---especially for rare or unseen categories---receive low confidence~\cite{zheng2024training,wang2024open}.

We address these issues by introducing \textbf{DetRefiner}, a lightweight, detector-agnostic plug-and-play module that \emph{post hoc} refines OVOD predictions.
DetRefiner operates purely on the outputs of existing detectors at inference time and jointly reasons about image-level and box-level confidence.
It produces semantically aligned and well-calibrated confidence scores for ambiguous, fine-grained, and rare or unseen categories.

Concretely, DetRefiner fuses global and local features extracted from DINO~\cite{caron2021emerging,simeoni2025dinov3} using a compact Transformer-based encoder~\cite{vaswani2017attention}, producing a \emph{class vector} that captures global scene semantics and a \emph{patch vector} that encodes local region evidence.
These embeddings jointly predict refined confidence scores, which are used to recalibrate the base detector's predictions and correct classification inconsistencies.
To further improve semantic alignment, calibration, and generalization to rare and unseen categories, we introduce a distillation-based alignment mechanism~\cite{hinton2015distilling,Romero15-iclr} that transfers semantic knowledge from CLIP-based vision--language models~\cite{radford2021learning,vasu2024mobileclip,zhang2024long}, encouraging both vectors to inherit strong vision--language alignment.
In practice, we run the base OVOD detector with a zero detection threshold and refine all candidate boxes in a single forward pass, enabling DetRefiner to rescue many low-scored but correct boxes while keeping the inference overhead moderate.
Figure~\ref{fig:lvis_detrefiner} and Figure~\ref{fig:qualitative_results} illustrate how our method improves detection results on LVIS by producing semantically coherent and well-calibrated outputs, while also addressing common failure modes.

Our contributions are threefold:
\begin{enumerate}
\item \textbf{Lightweight fusion of global and local features.}
We design a 7.5M-parameter Transformer encoder that fuses global and local features from DINO~\cite{caron2021emerging,simeoni2025dinov3} into class and patch vectors, which recalibrate base detector scores without modifying or retraining the detector.

\item \textbf{Distillation-enhanced calibration for rare and unseen categories.}
We apply a knowledge distillation loss from CLIP-based image features~\cite{radford2021learning,vasu2024mobileclip,zhang2024long} to both class and patch vectors, enabling DetRefiner to inherit strong vision--language alignment and improve calibration on rare and unseen classes.

\item \textbf{Consistent gains across diverse detectors and benchmarks.}
Our method consistently improves multiple OVOD models~\cite{li2022grounded,liu2024grounding,zhao2024open,fu2025llmdet} on COCO~\cite{lin2014microsoft}, LVIS~\cite{gupta2019lvis}, ODinW13~\cite{li2022grounded,li2022elevater}, and Pascal VOC~\cite{everingham2010pascal}, achieving up to \textbf{+10.1 AP} improvement on rare and unseen categories.
\end{enumerate}

\begin{figure*}[t]
  \centering
  \includegraphics[width=0.75\linewidth]{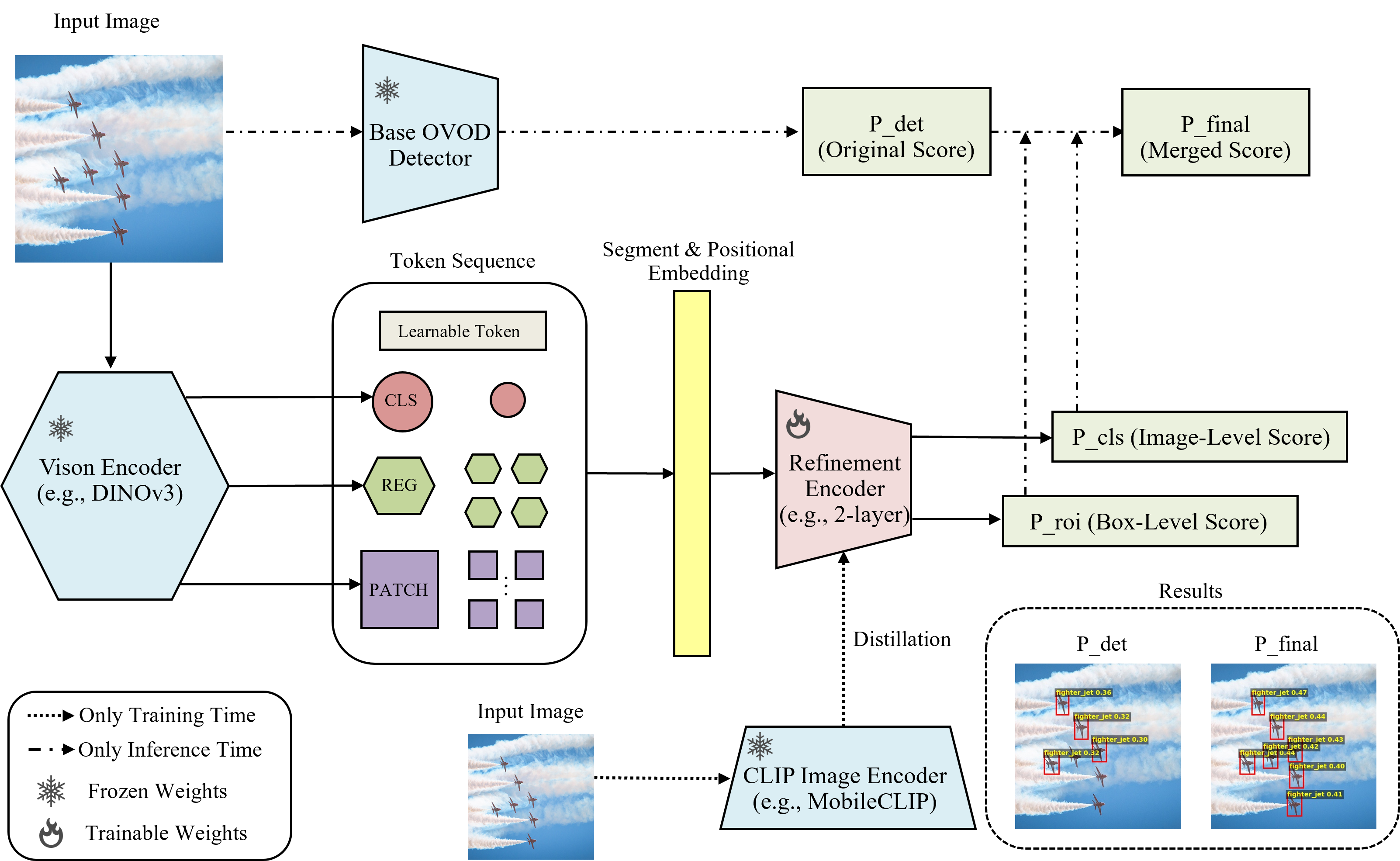}
  \caption{
\textbf{Overall pipeline of DetRefiner.} A base open-vocabulary detector takes an input image and outputs bounding boxes and category scores. In parallel, a vision encoder (e.g., DINOv3) extracts global and local features, which DetRefiner fuses and learns via classification and distillation losses. At inference, DetRefiner outputs image- and box-level confidence scores that are combined with the base detector scores to obtain the final detection confidence. For visualization of the results (bottom right), a box score threshold of 0.3 and an IoU threshold of 0.3 are applied for class-wise NMS.
  }
  \vspace{-0.6em}
  \label{fig:detrefiner_overview}
\end{figure*}

\section{Related Work}
\subsection{Open-Vocabulary Object Detection}
Open-vocabulary object detection (OVOD)~\cite{zareian2021open,kamath2021mdetr,guopen,zhong2022regionclip,du2022learning,minderer2022simple,zhou2022detecting,li2022grounded,liu2024grounding} extends traditional detection by recognizing categories beyond a fixed label set through vision--language alignment.
Early methods such as OVR-CNN~\cite{zareian2021open} and MDETR~\cite{kamath2021mdetr} ground textual queries to visual regions via multimodal pretraining, while ViLD~\cite{guopen} and RegionCLIP~\cite{zhong2022regionclip} align region proposals with textual embeddings for zero-shot recognition.
More recent models, including GLIP~\cite{li2022grounded} and Grounding DINO~\cite{liu2024grounding}, jointly optimize grounding and detection for stronger open-vocabulary generalization.

Our DetRefiner is complementary to these frameworks: instead of designing a new detector, it operates \emph{on top of} existing OVOD models and post hoc refines both image- and box-level confidence without modifying or retraining the base model.

\subsection{Complementary Approaches for OVOD Models}
To mitigate semantic misalignment and unreliable confidence estimation, recent studies propose complementary methods that augment existing detectors.
SIC-CADS~\cite{fang2023simple} performs image-level confidence calibration using contextual cues, improving reliability without altering the detector but leaving box-level scores unchanged.
DVDet~\cite{jin2024llms} uses LLM-assisted descriptor generation and conditional prompts to expand detector vocabularies and enhance semantic consistency, yet requires joint training with the base detector and access to internal features.
CODet~\cite{Li_2025_ICCV} leverages LLM-guided pseudo-labeling based on visual--textual co-occurrence to refine box-level predictions during training, and it also requires access to the detector's internal architecture.

In contrast, DetRefiner treats the base OVOD model as a black box while jointly refining image- and box-level confidence through lightweight semantic fusion.
Moreover, whereas DVDet~\cite{jin2024llms} and CODet~\cite{Li_2025_ICCV} mainly evaluate on earlier OVOD baselines~\cite{guopen,zhong2022regionclip,zhou2022detecting,wu2023aligning,linlearning,du2022learning}, we demonstrate DetRefiner’s practicality by attaching it to recent high-performing detectors~\cite{li2022grounded,liu2024grounding,zhao2024open,fu2025llmdet} across multiple benchmarks.
As some earlier OVOD baselines and complementary approaches are not publicly available or difficult to reproduce, we construct reproducible baselines following their published settings and evaluate DetRefiner on top of them for fair and reliable comparison.

\subsection{Fusion of Vision and Vision-Language Foundation Models}
Recent work explores combining complementary foundation models to improve visual--language understanding and transferability~\cite{kar2024brave,ranzinger2024radio,el2024probing,shieagle,tong2024eyes}.
BRAVE~\cite{kar2024brave} integrates features from multiple frozen vision encoders into unified representations, and AM-RADIO~\cite{ranzinger2024radio} distills knowledge from backbones such as CLIP~\cite{radford2021learning}, DINOv2~\cite{caron2021emerging}, and SAM~\cite{kirillov2023segment} into a generalist model, showing that combining CLIP’s global semantics with DINO’s rich local and global features improves robustness.

DetRefiner adopts a related philosophy but applies foundation-model complementarity to OVOD as a \emph{post-hoc} refinement module.
\section{Method}
\label{sec:method}

We propose \textbf{DetRefiner} (Figure~\ref{fig:detrefiner_overview}), a lightweight refinement module that operates on top of an arbitrary open-vocabulary detector.
Given an input image, we first extract rich visual representations from a frozen DINOv3 ~\cite{simeoni2025dinov3, darcet2024vision}.
A compact Transformer encoder then refines these features into an image-level \emph{class vector} and region-level \emph{patch vectors}, which are aligned with text embeddings from MobileCLIP~\cite{vasu2024mobileclip} through joint classification and distillation objectives.
At inference time, DetRefiner takes the base detector’s predictions---a fixed set of bounding boxes and their category-wise confidence scores---and keeps both the box coordinates and category labels unchanged, while refining the confidence scores for each \emph{(box, category)} pair.

\subsection{Preliminaries}

Let $\mathcal{I} \in \mathbb{R}^{H\times W\times 3}$ denote an input image.
We employ the DINOv3 Vision Transformer as a frozen feature extractor to obtain three levels of visual representation:
a global feature token $\mathbf{g} \in \mathbb{R}^{d_g}$, four register tokens $\{\mathbf{r}_i\}_{i=1}^{4} \in \mathbb{R}^{d_r}$, and patch features $\{\mathbf{p}_j\}_{j=1}^{196} \in \mathbb{R}^{d_p}$.
These features encode complementary contextual information---global semantics, intermediate structural priors, and fine-grained local cues, respectively.

Each feature type is independently projected into a common latent space through learnable linear transformations:
\begin{equation}
\mathbf{g}' = W_g \mathbf{g}, \quad
\mathbf{r}'_i = W_r \mathbf{r}_i, \quad
\mathbf{p}'_j = W_p \mathbf{p}_j,
\end{equation}
where $W_g$, $W_r$, and $W_p$ are linear layers mapping to a unified hidden dimension $d$.
A learnable class token $\mathbf{c}'$ is prepended to the input sequence to aggregate global semantics during self-attention.

To distinguish heterogeneous token types, we introduce learnable segment embeddings~\cite{kim2021vilt}, which assign a unique embedding to each token type (class, global, register, and patch) and are added to the corresponding token features. 
For spatial encoding, we apply fixed 2D sine-cosine positional embeddings to patch tokens, constructed based on the spatial grid of patch features following MAE-style designs~\cite{he2022masked}.
The complete token sequence is then formed as:
\begin{equation}
T = [\mathbf{c}^{\prime\prime}; \mathbf{g}^{\prime\prime}; \mathbf{r}^{\prime\prime}_1, \ldots, \mathbf{r}^{\prime\prime}_4; \mathbf{p}^{\prime\prime}_1, \ldots, \mathbf{p}^{\prime\prime}_{196}],
\end{equation}
which is processed by a lightweight Transformer encoder, referred to as the Refinement Encoder (Figure \ref{fig:detrefiner_overview}).

The Transformer output consists of:
\begin{itemize}
    \item a \textbf{class vector} $\mathbf{v}_{cls} \in \mathbb{R}^d$ (the output at the class token), representing image-level semantic content, and
    \item \textbf{patch vectors} $\{\mathbf{v}_{patch,j}\}_{j=1}^{196} \in \mathbb{R}^d$, encoding localized visual evidence.
\end{itemize}

For region-level reasoning, we further pool patch vectors within each bounding box by ROI Align~\cite{he2017mask} to obtain an \textbf{ROI patch vector} $\mathbf{v}_{roi}$.
This vector captures fine-grained appearance within predicted object regions, enabling region-wise calibration and classification. ROI Align pools patch features into region-level representations that suppress background context, while global features provide complementary scene-level priors. This design is important, as local cues alone can be visually and semantically ambiguous.
In training, bounding boxes and category labels are provided by the dataset annotations; at inference, we use the boxes from the base detector.

\subsection{Feature-Text Alignment Framework}
\label{sec:framework}

DetRefiner aligns visual and textual representations through joint optimization that enforces both global and local semantic consistency.
Given visual embeddings from DINOv3 and text embeddings from MobileCLIP, the model learns to associate global image semantics and region-level evidence with corresponding textual categories.

Let $\mathbf{v}_{cls}$ and $\mathbf{v}_{roi}$ denote the class-level and ROI-level visual embeddings, respectively.
We obtain normalized features
$\hat{\mathbf{v}}_{cls} = \mathbf{v}_{cls} / \|\mathbf{v}_{cls}\|_2$,
$\hat{\mathbf{v}}_{roi} = \mathbf{v}_{roi} / \|\mathbf{v}_{roi}\|_2$,
and normalized text embeddings $\{\hat{\mathbf{t}}_k\}_{k=1}^{C}$ for $C$ seen categories.
Their cosine similarities are converted into logits as:
\begin{equation}
\text{logit}_{cls,k} = s_{cls} \, (\hat{\mathbf{v}}_{cls}^\top \hat{\mathbf{t}}_k), 
\quad
\text{logit}_{roi,k} = s_{cls} \, (\hat{\mathbf{v}}_{roi}^\top \hat{\mathbf{t}}_k),
\end{equation}
where $s_{\mathrm{cls}} = 1 / \tau$ is a temperature scaling factor.

Training uses images that may contain both seen and unseen categories, but supervision is applied only to the seen label set: all losses are computed over $C_{seen}$ and ignore unseen labels.
For LVIS, we further treat only negatives from "neg\_category\_ids"\footnote{\url{https://www.lvisdataset.org/dataset}} as valid negative labels and ignore all other negatives when computing losses. We then optimize the model using the following losses:

\paragraph{Image-level classification loss.}
The class vector $\mathbf{v}_{cls}$ predicts the global presence of categories within an image using binary cross-entropy:
\begin{equation}
\begin{split}
\mathcal{L}^{cls}_{img}
= -\frac{1}{|C_{seen}|}
\sum_{k \in C_{seen}} \Big[
y_k \log \sigma(\text{logit}_{cls,k})
\\[-2pt]
+ (1 - y_k)
\log\!\big(1 - \sigma(\text{logit}_{cls,k})\big)
\Big],
\end{split}
\end{equation}
where $\sigma(\cdot)$ denotes the sigmoid function and $y_k \in \{0,1\}$ indicates the presence of category $k$ in the image.

\paragraph{Region-level classification loss.}
ROI patch embeddings $\mathbf{v}_{roi}$, pooled from region proposals via ROI Align, are supervised similarly to predict the object category in each bounding box:
\begin{equation}
\begin{split}
\small
\mathcal{L}^{cls}_{roi}
= -\frac{1}{|B||C_{seen}|}
\sum_{b=1}^{B}
\sum_{k \in C_{seen}} \Big[
y_k \log \sigma(\text{logit}_{roi,k}) 
\\[-2pt]
+ (1 - y_k)
\log\!\big(1 - \sigma(\text{logit}_{roi,k})\big)
\Big],
\end{split}
\end{equation}
where $y_k$ is defined according to the ground-truth label of the region and $B$ denotes the number of ROIs for an image.

\paragraph{Global distillation loss.}
To transfer semantic alignment from MobileCLIP, we apply a cosine similarity constraint between the learned class vector and the CLIP-aligned image feature $\mathbf{m}_{img}$:
\begin{equation}
\mathcal{L}_{ckd}
= 1 - 
\frac{
\mathbf{v}_{cls}^\top \mathbf{m}_{img}
}{
\|\mathbf{v}_{cls}\|_2 \, \|\mathbf{m}_{img}\|_2
}.
\end{equation}
This loss encourages the global representation of DetRefiner to inherit the semantic structure of the vision--language teacher at the image level.

\paragraph{Local distillation loss.}
To encourage locally aggregated representations to share the same semantic direction as the CLIP-aligned visual space, we align the average-pooled patch vectors with the MobileCLIP image feature. 
Let $\bar{\mathbf{v}}_{patch} = \frac{1}{196} \sum_{j=1}^{196} \mathbf{v}_{patch,j}$ denote the mean of the patch embeddings.
The local distillation loss is then defined as:
\begin{equation}
\mathcal{L}_{pkd}
= 1 - 
\frac{
\bar{\mathbf{v}}_{patch}^\top \mathbf{m}_{img}
}{
\|\bar{\mathbf{v}}_{patch}\|_2 \, \|\mathbf{m}_{img}\|_2
}.
\end{equation}
This cosine similarity loss aligns the pooled patch representation with the CLIP-aligned image feature, encouraging DetRefiner's locally aggregated features to remain consistent with the global semantic structure of the vision--language teacher.

\paragraph{Overall objective.}
The overall objective combines classification and distillation losses, each serving complementary roles in enforcing semantic alignment and confidence calibration:

\begin{equation}
\mathcal{L} = 
\mathcal{L}^{cls}_{img} +
\mathcal{L}^{cls}_{roi} +
\lambda_1 \mathcal{L}_{ckd} +
\lambda_2 \mathcal{L}_{pkd},
\end{equation}
where $\lambda_1$ and $\lambda_2$ are the weights for the distillation terms.

\subsection{Inference}
Given a novel image, the base OVOD model produces bounding boxes and class scores for a set of candidate categories.
In parallel, MobileCLIP provides text embeddings $\{\mathbf{t}_k\}$ for these categories, and DetRefiner computes similarity-based probabilities for both class and ROI patch vectors:
\begin{equation}
P_{cls,k} = \sigma\!\big(s_{cls} \, (\hat{\mathbf{v}}_{cls}^\top \hat{\mathbf{t}}_k)\big), \quad
P_{roi,k} = \sigma\!\big(s_{cls} \, (\hat{\mathbf{v}}_{roi}^\top \hat{\mathbf{t}}_k)\big).
\end{equation}

The final detection confidence is obtained by linearly combining the base detector score $P_{det}$ with these semantic cues. $(w_d, w_c, w_p)$ are the weights parameters:
\begin{equation}
P_{final} = w_d P_{det} + w_c P_{cls} + w_p P_{roi},
\end{equation}
Importantly, DetRefiner does not alter box coordinates or category assignments, but only recalibrates confidence scores.

Through this mechanism, unseen categories---defined solely by text embeddings at inference---can be recognized in a zero-shot manner.
The class vector captures global scene priors, while ROI patch vectors refine fine-grained localization, jointly improving open-vocabulary detection performance.

\begin{figure*}[t]
  \centering
  \includegraphics[width=1.0\linewidth]{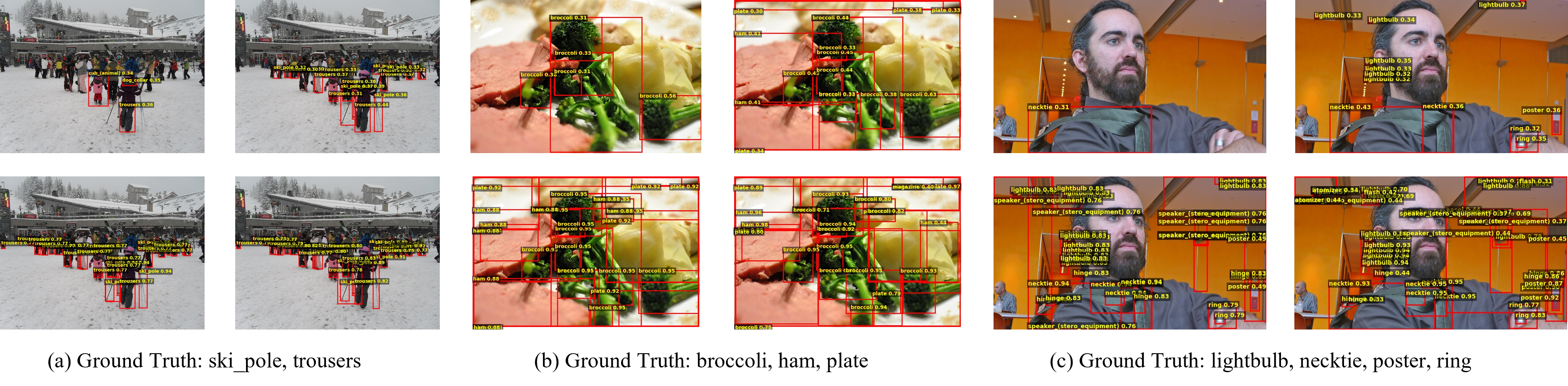}
  \caption{
Qualitative comparison of detection results before and after applying DetRefiner.
Top: base detector (left) vs. base detector + DetRefiner (right).
Bottom: predictions based on the class vector (left) and patch vector (right).
DetRefiner suppresses overconfident false positives and recovers missed objects by combining global and local cues. For visualization, a box score threshold of 0.3 and an IoU threshold of 0.3 are applied for class-wise NMS on all images. More visualization results are in Appendix 2.
  }
  \vspace{-0.6em}
  \label{fig:qualitative_results}
\end{figure*}

\begin{table*}[t]
\centering
\setlength{\tabcolsep}{4pt}
\caption{
Open-vocabulary detection results on \textbf{OV-LVIS}~\cite{gupta2019lvis} with and without DetRefiner.
AP denotes $\mathrm{AP}@[0.5{:}0.95]$. MM-Grounding-DINO-Large~\cite{zhao2024open} shows lower performance likely due to training data differences (e.g., V3Det~\cite{wang2023v3det}); similar trends are observed in prior work (e.g., LLMDet~\cite{fu2025llmdet}).
}
\vspace{-6pt}
\scriptsize
\begin{tabular}{@{}l l cccc@{}}
\toprule
Method & Trained Data & AP$_r$ & AP$_c$ & AP$_f$ & AP$_{all}$ \\
\midrule
GLIP (Tiny)~\cite{li2022grounded}  & O365, GoldG, CC3M, SBU & 18.0 & 20.5 & 32.3 & 26.0 \\
\hspace{3mm}+DetRefiner & +OV-LVIS & 22.7\gain{+4.7} & 25.2\gain{+4.7} & 33.8\gain{+1.5} & 29.1\gain{+3.1} \\
\midrule
GLIP (Large)~\cite{li2022grounded}  & FourODs, GoldG, CC3M+12M, SBU & 29.1 & 34.7 & 42.3 & 37.9 \\
\hspace{3mm}+DetRefiner & +OV-LVIS & 30.2\gain{+1.1} & 37.6\gain{+2.9} & 42.9\gain{+0.6} & 39.5\gain{+1.6} \\
\midrule
Grounding DINO (Tiny)~\cite{liu2024grounding}  & O365, GoldG, Cap4M & 19.9 & 22.6 & 33.1 & 27.4 \\
\hspace{3mm}+DetRefiner & +OV-LVIS & \best{30.0}\gain{+10.1} & 31.6\gain{+9.0} & 37.1\gain{+4.0} & 34.1\gain{+6.7} \\
\midrule
Grounding DINO (Base)~\cite{liu2024grounding} & \makecell[l]{COCO, O365, GoldG, Cap4M,\\ OpenImages, ODinW35, RefCOCO} & 27.3 & 31.6 & 36.5 & 33.6 \\
\hspace{3mm}+DetRefiner & +OV-LVIS & 33.5\gain{+6.2} & 37.1\gain{+5.5} & 37.0\gain{+0.5} & 36.7\gain{+3.1} \\
\midrule
MM-Grounding DINO (Tiny)~\cite{zhao2024open} & O365, GoldG, GRIT, V3Det & 34.3 & 35.7 & 45.9 & 40.5 \\
\hspace{3mm}+DetRefiner & +OV-LVIS & 39.4\gain{+5.1} & 40.7\gain{+5.0} & 48.1\gain{+2.2} & 44.2\gain{+3.7} \\
\midrule
MM-Grounding DINO (Base)~\cite{zhao2024open}  & O365, GoldG, V3Det & 35.4 & 37.8 & 48.5 & 42.8 \\
\hspace{3mm}+DetRefiner & +OV-LVIS & 41.6\gain{+6.2} & 43.0\gain{+5.2} & 50.4\gain{+1.9} & 46.4\gain{+3.6} \\
\midrule
MM-Grounding DINO (Large)~\cite{zhao2024open}  & O365V2, OpenImageV6, GoldG & 30.4 & 32.0 & 42.7 & 37.0 \\
\hspace{3mm}+DetRefiner & +OV-LVIS & 34.2\gain{+3.8} & 38.2\gain{+6.2} & 43.5\gain{+0.8} & 40.4\gain{+3.4} \\
\midrule
LLMDet (Tiny)~\cite{fu2025llmdet} & GroundingCap-1M & 36.7 & 37.6 & 49.7 & 43.4 \\
\hspace{3mm}+DetRefiner & +OV-LVIS & 41.9\gain{+5.2} & 44.5\gain{+6.9} & 51.9\gain{+2.2} & 47.9\gain{+4.5} \\
\midrule
LLMDet (Base)~\cite{fu2025llmdet}  & GroundingCap-1M & 39.2 & 42.1 & 53.1 & 47.2 \\
\hspace{3mm}+DetRefiner & +OV-LVIS & 46.1\gain{+6.9} & 47.3\gain{+5.2} & 55.0\gain{+1.9} & 50.9\gain{+3.7} \\
\midrule
LLMDet (Large)~\cite{fu2025llmdet}  & GroundingCap-1M & 43.9 & 43.6 & 55.4 & 49.3 \\
\hspace{3mm}+DetRefiner & +OV-LVIS & 50.1\gain{+6.2} & 49.3\gain{+5.7} & 57.1\gain{+1.7} & 53.2\gain{+3.9} \\
\midrule
\end{tabular}
\label{tab:ov-lvis_results}
\vspace{-0.6em}
\end{table*}

\begin{table}[t]
\centering
\setlength{\tabcolsep}{4pt}
\caption{
Open-vocabulary detection results on \textbf{OV-COCO}~\cite{lin2014microsoft} using baseline detectors trained with the same data as in Table \ref{tab:ov-lvis_results}, and a DetRefiner trained on OV-COCO, evaluated with and without DetRefiner.
AP denotes $\mathrm{AP}@[0.5{:}0.95]$.
}
\vspace{-6pt}
\scriptsize
\begin{tabular}{@{}l cccc@{}}
\toprule
Method & AP$^{novel}$ & AP$^{base}$ & AP$^{all}$ \\
\midrule
GLIP (Tiny)~\cite{li2022grounded} & 53.4 & 44.2 & 46.6 \\
\hspace{3mm}+DetRefiner & 53.8\gain{+0.4} & 46.3\gain{+2.1} & 48.2\gain{+1.6} \\
\midrule
Grounding DINO (Tiny)~\cite{liu2024grounding} & 57.4 & 46.6 & 49.4 \\
\hspace{3mm}+DetRefiner & \best{58.1}\gain{+0.7} & 48.9\gain{+2.3} & 51.3\gain{+1.9} \\
\midrule
MM-Grounding DINO (Tiny)~\cite{zhao2024open} & 59.3 & 47.8 & 50.8 \\
\hspace{3mm}+DetRefiner & 59.5\gain{+0.2} & 49.4\gain{+1.6} & 52.1\gain{+1.3} \\
\midrule
LLMDet (Tiny)~\cite{fu2025llmdet} & 61.6 & 53.0 & 55.3 \\
\hspace{3mm}+DetRefiner & 61.7\gain{+0.1} & 54.0\gain{+1.0} & 56.0\gain{+0.7} \\
\midrule
\end{tabular}
\label{tab:ov-coco_results_summary}
\vspace{-0.6em}
\end{table}

\begin{table}[t]
\centering
\setlength{\tabcolsep}{4pt}
\caption{
Cross-dataset generalization on \textbf{ODinW13}~\cite{li2022grounded,li2022elevater} and \textbf{Pascal VOC}~\cite{everingham2010pascal}  using baseline detectors trained with the same data as in Table \ref{tab:ov-lvis_results}, and a DetRefiner trained on OV-LVIS, evaluated with and without DetRefiner.
AP denotes $\mathrm{AP}@[0.5{:}0.95]$. 
}
\vspace{-6pt}
\scriptsize
\begin{tabular}{@{}l cccc@{}}
\toprule
Method & mAP (ODinW13) & AP (Pascal VOC)\\
\midrule
GLIP (Tiny)~\cite{li2022grounded} & 41.7 & 56.6 \\
\hspace{3mm}+DetRefiner & \best{43.3}\gain{+1.6} & 59.2\gain{+2.6} \\
\midrule
Grounding DINO (Tiny)~\cite{liu2024grounding} & 38.4 & 56.3 \\
\hspace{3mm}+DetRefiner & 38.9\gain{+0.5} & \best{60.0}\gain{+3.7} \\
\midrule
MM-Grounding DINO (Tiny)~\cite{zhao2024open} & 41.1 & 57.6 \\
\hspace{3mm}+DetRefiner & 42.2\gain{+1.1} & 60.4\gain{+2.8} \\
\midrule
LLMDet (Tiny)~\cite{fu2025llmdet} & 31.5 & 59.4 \\
\hspace{3mm}+DetRefiner & 32.2\gain{+0.7} & 61.3\gain{+1.9} \\
\midrule
\end{tabular}
\label{tab:odinw13_pascal_results_summary}
\vspace{-0.6em}
\end{table}

\begin{table}[t]
\centering
\footnotesize
\setlength{\tabcolsep}{1.5pt}
\renewcommand{\arraystretch}{0.9}
\caption{
Ablation on main components for DetRefiner using Grounding DINO (Tiny). ``Refine'' indicates the presence of Refinement Encoder. 
$L_{ckd}$ and $L_{pkd}$ denote distillation on class and patch vectors. ``CLIP-V'' indicates direct CLIP visual features fed into the refinement encoder.
}
\vspace{-6pt}
\scriptsize
\begin{tabular}{cccc|cccc|ccc}
\toprule
\multicolumn{4}{c|}{\textbf{Main Components}} &
\multicolumn{4}{c|}{\textbf{OV-LVIS}} &
\multicolumn{3}{c}{\textbf{OV-COCO}} \\
\midrule
Refine & $L_{ckd}$ & $L_{pkd}$ & CLIP-V & AP$_r$ & AP$_c$ & AP$_f$ & AP$_{all}$ & AP$^{novel}$ & AP$^{base}$ & AP$^{all}$ \\
\midrule
\checkmark & \checkmark & \checkmark &  & \best{30.0} & 31.6 & 37.1 & 34.1 & \textbf{58.1} & 48.9 & 51.3 \\
\midrule
 & \checkmark & \checkmark & & 26.9 & 29.8 & 37.0 & 33.1 & 57.7 & 48.1 & 50.6 \\
\checkmark & \checkmark & & & 29.1 & 31.6 & 37.1 & 34.1 & 58.1 & 49.0 & 51.3 \\
\checkmark &  & \checkmark &  & 29.2 & 31.4 & 37.2 & 34.0 & 56.9 & 49.0 & 51.0 \\
\checkmark &  & & & 26.4 & 31.2 & 37.0 & 33.6 & 56.8 & 49.0 & 51.0\\
\checkmark &  &  & \checkmark & 26.1 & 31.6 & 37.1 & 33.8 & 56.9 & 49.0 & 51.0 \\
\bottomrule
\end{tabular}
\label{tab:kd_loss_inputs}
\end{table}

\begin{table*}[t]
\centering
\setlength{\tabcolsep}{4pt}
\caption{
Ablation on Transformer depth, temperature $\tau$, feature encoders, and ROI extraction in DetRefiner using Grounding DINO (Tiny).
Baseline: 2 layers, $\tau{=}0.03$, DINOv3, MobileCLIP, ROI Align.
Here, ''Inclusion'' denotes averaging all patch vectors whose corresponding patches overlap the ground-truth bounding box, and ''DINOv2-reg'' denotes DINOv2 features augmented with register tokens~\cite{darcet2024vision}.
}
\vspace{-6pt}
\scriptsize
\begin{tabular}{@{}cccccccccccc@{}}
\toprule
\textbf{num\_layers} & $\boldsymbol{\tau}$ & \textbf{DINO} & \textbf{CLIP} & \textbf{ROI} &
\multicolumn{4}{c}{\textbf{OV-LVIS}} & \multicolumn{3}{c}{\textbf{OV-COCO}} \\
\cmidrule(lr){6-9} \cmidrule(lr){10-12}
& & & & & AP$_r$ & AP$_c$ & AP$_f$ & AP$_{all}$ & AP$^{novel}$ & AP$^{base}$ & AP$^{all}$ \\
\midrule
\multicolumn{12}{l}{\textit{Baseline}} \\
2 & 0.03 & DINOv3 & MobileCLIP & ROI Align & \textbf{30.0} & 31.6 & 37.1 & 34.1 & \textbf{58.1} & 48.9 & 51.3 \\
\midrule
\multicolumn{12}{l}{\textit{Varying Transformer depth}} \\
1 &  &  &  &  & 29.1 & 31.7 & 37.2 & 34.1 & 57.9 & 48.8 & 51.2 \\
3 &  &  &  &  & 28.5 & 31.0 & 37.1 & 33.7 & 58.2 & 49.0 & 51.4 \\
6 &  &  &  &  & 25.5 & 27.6 & 36.2 & 31.6 & \best{58.3} & 49.0 & 51.4 \\
12 &  &  &  &  & 20.2 & 23.0 & 33.6 & 27.9 & 58.2 & 48.9 & 51.4 \\
\midrule
\multicolumn{12}{l}{\textit{Varying temperature $\tau$}} \\
 & 0.01 &  &  &  & 29.3 & 32.3 & 37.2 & 34.4 & 58.2 & 48.8 & 51.3 \\
 & 0.05 &  &  &  & 28.5 & 30.8 & 37.0 & 33.6 & 57.7 & 49.0 & 51.2 \\
 & 0.10 &  &  &  & 26.1 & 28.6 & 36.5 & 32.2 & 57.3 & 48.8 & 51.0 \\
\midrule
\multicolumn{12}{l}{\textit{Different feature encoders}} \\
 &  & DINOv2-reg & &  & 28.6 & 31.8 & 36.9 & 34.0 & 58.2 & 48.8 & 51.2 \\
 &  & & CLIP &  & 26.4 & 30.6 & 36.9 & 33.3 & 57.4 & 48.9 & 51.1 \\
 &  & & LongCLIP &  & 27.1 & 30.2 & 37.0 & 33.2 & 57.4 & 48.8 & 51.2 \\
\midrule
\multicolumn{12}{l}{\textit{ROI extraction methods}} \\
 &  &  & & Inclusion & 29.5 & 31.3 & 37.1 & 33.9 & 58.1 & 48.9 & 51.3 \\
 &  &  & & ROI Pooling & 29.0 & 31.1 & 37.0 & 33.8 & 58.1 & 48.8 & 51.2 \\
\bottomrule
\end{tabular}
\label{tab:ablation_hyperparam}
\vspace{-0.5em}
\end{table*}

\begin{table}[t]
\centering
\caption{
Ablation on ensemble weights $(w_d,w_c,w_p)$ on OV-LVIS and OV-COCO using Grounding DINO (Tiny).
}
\vspace{-6pt}
\scriptsize
\setlength{\tabcolsep}{2pt}
\begin{tabular}{ccc|cccc|ccc}
\toprule
\multicolumn{3}{c|}{\textbf{Ensemble Weight}} &
\multicolumn{4}{c|}{\textbf{OV-LVIS}} &
\multicolumn{3}{c}{\textbf{OV-COCO}} \\
\midrule
$w_d$ & $w_c$ & $w_p$ & AP$_r$ & AP$_c$ & AP$_f$ & AP$_{all}$ & AP$^{novel}$ & AP$^{base}$ & AP$^{all}$ \\
\midrule
0.8 & 0.10  & 0.10  & \best{30.0} & 31.6 & 37.1 & 34.1 & \textbf{58.1} & 48.9 & 51.3 \\
\midrule
0.5 & 0.25 & 0.25 & 30.2 & 35.2 & 35.4 & 34.9 & 54.5 & 46.4 & 48.5 \\
0.6 & 0.20  & 0.20  & 30.1 & 35.8 & 36.5 & 35.6 & 56.8 & 47.6 & 50.0 \\
0.7 & 0.15 & 0.15 & \textbf{30.6} & 34.6 & 37.2 & 35.5 & 57.8 & 48.6 & 51.0 \\
0.9 & 0.05 & 0.05 & 24.7 & 27.5 & 35.8 & 31.3 & 57.9 & 48.4 & 50.9 \\
\midrule
0.8 & 0.20 & 0   & 28.0 & 31.7 & 36.9 & 33.9 & \best{58.2} & 48.5 & 51.0 \\
0.8 & 0   & 0.20 & 29.1 & 30.8 & 37.0 & 33.7 & 56.9 & 48.0 & 50.8 \\
\bottomrule
\end{tabular}
\label{tab:ensemble_weights}
\end{table}

\begin{table}[t]
  \centering
\scriptsize
\caption{Inference cost of GDINO-T with and without DetRefiner. DetRefiner introduces moderate latency and VRAM overhead.}
\vspace{-6pt}
  \label{tab:inference_cost}
  \begin{tabular}{lccc}
    \toprule
    Model & Latency (ms)$\downarrow$ & FPS$\uparrow$ & GPU Mem. (MB)$\downarrow$ \\
    \midrule
    GDINO-T   & 324  & 3.08  & 4752  \\
    + DetRefiner & 428 & 2.01 & 6984 \\
    \midrule
\multicolumn{4}{l}{\textit{Components of DetRefiner}} \\
    DINOv3 & 75 & 13.3 & 0 (CPU) \\
    Refinement Encoder & 29 & 34.5 & 2232 \\
    \bottomrule
  \end{tabular}
\end{table}

\section{Experiments}

\subsection{Datasets}
We evaluate DetRefiner on four benchmarks, covering both zero-shot and cross-dataset settings.

\noindent\textbf{OV-COCO.}
Following OVR-CNN~\cite{zareian2021open}, we split COCO~\cite{lin2014microsoft} into 48 base classes (seen) and 17 novel classes (unseen).
The training set contains 118{,}287 images and the validation set includes 5{,}000 images.
We use COCO API\footnote{\url{https://github.com/cocodataset/cocoapi}} for evaluation.

\noindent\textbf{OV-LVIS.}
Following ViLD~\cite{guopen}, we use LVIS~\cite{gupta2019lvis} with 405 frequent and 461 common classes as seen, and 337 rare classes as unseen.
We adopt the \textit{minival} split~\cite{li2022grounded}, consisting of 100{,}170 training images and 4{,}809 validation images.
We use LVIS API\footnote{\url{https://github.com/lvis-dataset/lvis-api}} for evaluation.

\noindent\textbf{ODinW13.}
To assess cross-dataset generalization, we follow~\cite{li2022grounded,li2022elevater} and evaluate OV-LVIS-trained detectors on the ODinW13 suite.
Table 1 in Appendix 1 summarizes the number of classes and test images for each dataset.

\noindent\textbf{Pascal VOC.}
Because several ODinW13 datasets contain very few test images and classes, we additionally report results on Pascal VOC~\cite{everingham2010pascal}, a subset of ODinW13 with 20 categories and over 3{,}000 test images.

\subsection{Implementation Details}
Global and local visual features extracted from DINOv3 (ViT-B/16)~\cite{simeoni2025dinov3} are used as inputs to Refinement Encoder, while MobileCLIP-B~\cite{vasu2024mobileclip} provides global visual features for distillation and text embeddings for classification. Text prompts are constructed using raw category names without templates.
All image and text features are pre-extracted and fixed during training, and no data augmentation is applied.

The refinement encoder is a 2-layer Transformer (hidden size 512, 8 heads)~\cite{vaswani2017attention}. 

We train with AdamW~\cite{KingmaB14,loshchilovdecoupled} for 30 epochs (batch size 64), using a learning rate of $1\times10^{-3}$, weight decay 0.01, cosine learning rate decay, and EMA~\cite{tarvainen2017mean,izmailov2018averaging} for stability.
We set the loss weights for the distillation terms to $\lambda_1 = \lambda_2 = 0.1$, and use a temperature of $\tau = 0.03$ with a label-smoothing ratio of $0.2$.

\subsection{Evaluation Protocol}

Unless otherwise noted, we train a DetRefiner on OV-LVIS and evaluate it on OV-LVIS, ODinW13, and Pascal VOC. For OV-COCO experiments, we train a separate DetRefiner on OV-COCO.

Evaluations on OV-COCO and OV-LVIS are designed to assess in-domain performance under an open-vocabulary setting, where the model is tested on novel categories beyond those seen during training. This allows us to isolate the model’s ability to generalize across label space while keeping the data distribution consistent.

In contrast, evaluation on ODinW13 is intended to measure out-of-domain generalization under distribution shift. As some categories overlap with the training data, this setting is not strictly zero-shot, but provides a complementary perspective on robustness to domain changes.

\subsection{Main Results}
We reproduce ten representative open-vocabulary detectors under a unified evaluation protocol, and report all improvements based on these baselines for fair comparison. Details are in Appendix 1.

\subsubsection{Results on OV-LVIS}
Table~\ref{tab:ov-lvis_results} summarizes OV-LVIS results, where DetRefiner consistently improves all base detectors across diverse backbones and training data.
Gains are especially pronounced on rare classes (e.g., +10.1 AP$_r$ for Grounding DINO (Tiny)), indicating that post-hoc semantic refinement is particularly beneficial for long-tail categories.
Larger backbones also improve across AP$_r$, AP$_c$, AP$_f$, and AP$_{all}$, showing that DetRefiner complements strong OVOD baselines by enhancing alignment and calibration.

\subsubsection{Results on OV-COCO}
Table~\ref{tab:ov-coco_results_summary} reports results on OV-COCO using Tiny-scale detectors.
DetRefiner yields consistent gains for all models, with the largest improvement of +1.9 AP$^{all}$ on Grounding DINO (Tiny).
Improvements on AP$^{novel}$ are modest but positive, as COCO contains fewer rare or fine-grained categories and thus is less challenging than LVIS. These results confirm that DetRefiner provides robust benefits across datasets, while the magnitude of gains correlates with the difficulty of semantic calibration.

\subsubsection{Results on ODinW13 and Pascal VOC} Table~\ref{tab:odinw13_pascal_results_summary} shows cross-dataset generalization results on ODinW13 and Pascal VOC when all detectors are trained on OV-LVIS and evaluated zero-shot. DetRefiner provides consistent gains across both datasets and all models, indicating that the refined confidence scores transfer well under domain shift. The largest gain is +3.7 AP on Pascal VOC with Grounding DINO (Tiny). Since Pascal VOC provides a larger and more balanced test set than most ODinW13 datasets, these improvements demonstrate robustness on more stable cross-dataset evaluations.

\subsection{Ablation Studies}
We now ablate components of DetRefiner, analyzing the Refinement Encoder and CLIP-based distillation, architectural and hyperparameter choices, and ensemble weights for combining detector and refinement scores.

\subsubsection{Effect of Main Components}
Table~\ref{tab:kd_loss_inputs} analyzes the contribution of the refinement encoder, distillation losses ($L_{ckd}$ and $L_{pkd}$), and direct CLIP feature inputs.
To isolate the source of gains, we compare with a projector-only baseline that uses identical features, losses, and training data, but removes the refinement encoder. The resulting performance drop indicates that explicit global–local interaction plays a critical role, supporting the effectiveness of the detect-then-refine design beyond stronger features or supervision alone.
Using both class-level ($L_{ckd}$) and patch-level ($L_{pkd}$) distillation yields the best performance, demonstrating that global and local alignment signals are complementary.
We use CLIP global visual features via distillation rather than as direct inputs, as direct CLIP inputs introduce a distribution mismatch with DINOv3 features. Since CLIP visual and text embeddings share the same space, jointly aligning refiner outputs to both provides a more stable and optimization-friendly objective.

\subsubsection{Model Architecture and Hyperparameters}
Table~\ref{tab:ablation_hyperparam} studies the effect of Transformer depth, temperature $\tau$, feature encoders, and ROI extraction.
A shallow 2-layer Transformer with $\tau{=}0.03$, DINOv3, MobileCLIP, and ROI Align provides the best trade-off between accuracy and complexity; deeper models or higher temperatures lead to degradation, likely due to overfitting or oversmoothing of confidence scores.
Replacing DINOv3 or MobileCLIP with alternative encoders slightly lowers AP, indicating the importance of the visual backbone. Nevertheless, our method consistently improves performance even with alternative backbones such as DINOv2 or CLIP, demonstrating robustness to backbone variations. Variations in ROI extraction have a smaller but noticeable impact. We adopt DINOv3 as the visual backbone rather than MobileCLIP, as MobileCLIP is optimized for global image–text alignment, whereas DINO-style features provide fine-grained, spatially localized representations better suited for detection.

\subsubsection{Ensemble Weights}
Table~\ref{tab:ensemble_weights} evaluates different ensemble weights $(w_d,w_c,w_p)$ for combining the base detector score with class- and ROI-level scores from DetRefiner using Grounding DINO (Tiny).
The default configuration $(0.8, 0.1, 0.1)$ yields stable gains on both OV-LVIS and OV-COCO and is used in all main experiments. 
In OVOD, many predictions have competing confidence scores, and even a modest contribution from DetRefiner (e.g., 20\%) can meaningfully affect ranking and overall performance.
Giving too little weight to the detector score or over-emphasizing DetRefiner tends to hurt performance, confirming that DetRefiner works best as a calibration-aware refinement module rather than a standalone detector.
While learning ensemble weights could further improve performance, it would require tighter integration with the base detector, which is beyond the scope of our post-hoc, model-agnostic design.

\subsection{Training and Inference Cost}
Table~\ref{tab:inference_cost} reports the inference cost of Grounding DINO-Tiny with and without DetRefiner.
Latency and VRAM are measured on an RTX~2080Ti and an Intel(R) Core(TM) i9-9900K CPU @ 3.60GHz.
In our setup, DINOv3 feature extraction runs on the CPU; moving it to the GPU would reduce latency at the cost of higher VRAM usage.
DetRefiner processes all proposals in a single forward pass of the refinement encoder, so even when the number of boxes is large, the additional latency remains moderate.\\
From the training perspective, DetRefiner is lightweight: all experiments are conducted on a single NVIDIA V100 GPU for 30 epochs (about 15 hours on OV-COCO or 20 hours on OV-LVIS), after which each DetRefiner can be attached to any detector without further fine-tuning.

\subsection{Visualization}
Figure~\ref{fig:qualitative_results} illustrates how DetRefiner refines predictions from the base detector. Compared to the original outputs (top-left), DetRefiner (top-right) suppresses overconfident false positives and recovers missed objects using both global and local evidence. The bottom row shows the contributions of class and patch vectors: the former captures scene-level context, while the latter emphasizes region-level details, jointly improving open-vocabulary detection reliability.
\section{Conclusion}
We present DetRefiner, a lightweight drop-in module for open-vocabulary object detection (OVOD). It improves detection by unifying global and local features from foundation models such as DINOv3~\cite{simeoni2025dinov3} and processing them with a compact Transformer encoder to produce image- and patch-level confidence scores for refinement. Unlike conventional approaches, DetRefiner operates solely on detection outputs, requiring no access to internal features or retraining of the base OVOD model.
While DetRefiner cannot recover objects that are entirely missed by the base detector, using a zero detection threshold helps rescue many low-scored but correct boxes in practice. \\
Thanks to its model-agnostic design, DetRefiner can also be applied to black-box detectors such as T-Rex2~\cite{jiang2024t} and Grounding DINO 1.5~\cite{ren2024grounding}, which are not publicly available; we plan to explore their use via APIs in future work.
{
    \small
    \bibliographystyle{ieeenat_fullname}
    \bibliography{main}
}

\clearpage
\setcounter{page}{1}
\maketitlesupplementary
\setcounter{section}{0}

\section{Reproducible Evaluation of Open-Vocabulary Object Detectors}
\label{sec:reproduction}

Before applying DetRefiner, we reproduce ten representative open-vocabulary detectors across all benchmarks under a unified evaluation protocol. All reported improvements are measured with respect to these reproduced baselines to ensure consistent comparison.

GLIP~\cite{li2022grounded} is implemented using its official repository\footnote{\url{https://github.com/microsoft/GLIP/tree/main}} with \texttt{min\_image\_size} set to 800. 
All other detectors are instantiated via the HuggingFace \textit{transformers} library\footnote{\url{https://huggingface.co/docs/transformers/model_doc/grounding-dino}}\footnote{\url{https://huggingface.co/docs/transformers/model_doc/mm-grounding-dino}}\footnote{\url{https://huggingface.co/collections/iSEE-Laboratory/llmdet}}.

For evaluation, we set the box score threshold to 0 for all models, ensuring that all predicted boxes are retained before computing AP. 
Unless otherwise specified, we use default inference-time settings for each model (e.g., NMS thresholds), except for the unified zero score threshold and the GLIP image resolution setting.

We note that differences in implementation (e.g., HuggingFace vs. official repositories), evaluation configurations such as the maximum number of predictions per image (300 for COCO and 100{,}000 for LVIS in our setup), and dataset-specific choices (e.g., ODinW13 test splits) may lead to discrepancies with originally reported results. 
To facilitate reproducibility, we release the full evaluation code and configurations at \url{https://github.com/hitachi-rd-cv/detrefiner}.

For \textbf{COCO}~\cite{lin2014microsoft}, we follow the common open-vocabulary setting and feed a single concatenated text prompt containing all 80 category names into models. The category names are concatenated in ascending order of the category IDs defined in the ground-truth annotations.

For \textbf{LVIS}~\cite{gupta2019lvis}, we adopt the same strategy as GLIP~\cite{li2022grounded}: the 1{,}203 category names are first sorted by their category IDs and then partitioned into groups of 40, where each group is used as a separate text prompt. The last three categories, which do not form a full group of 40, are used as a single text prompt containing only these three category names.

For both COCO and LVIS, we apply a simple text pre-processing step to the category names: we convert all characters to lowercase, replace underscores and hyphens with spaces, and remove parentheses. We observed that these pre-processing steps have negligible impact on the final performance, but we apply them consistently for reproducibility.

For \textbf{ODinW13}~\cite{li2022grounded,li2022elevater}, we follow the official GLIP configuration\footnote{\url{https://github.com/microsoft/GLIP/tree/main/configs}}. We use as test images the datasets specified in the \texttt{DATASETS:TEST} field of each YAML file in the configuration directory, and we construct the text prompts from the corresponding \texttt{OVERRIDE\_CATEGORY} field. The number of categories and test images for each ODinW13 dataset is summarized in Table~\ref{tab:odinw13_stats}.

\begin{table}[t]
  \centering
  \caption{Detailed statistics of the ODinW13 datasets~\cite{li2022grounded,li2022elevater}.}
  \scriptsize
  \resizebox{\linewidth}{!}{
      \begin{tabular}{lccc}
        \hline
        Dataset & \#Classes & Test Images \\
        \hline
        AerialMaritimeDrone-Large & 5 & 15 \\
        Aquarium-Combined & 7 & 127 \\
        CottontailRabbits & 1 & 19 \\
        EgoHands-Generic & 1 & 200 \\
        North-American-Mushrooms & 2 & 5 \\
        Packages-Raw & 1 & 4 \\
        \textbf{PascalVOC} & \textbf{20} & \textbf{3422} \\
        Pistols-Export & 1 & 297 \\
        Pothole & 1 & 133 \\
        Raccoon & 1 & 29 \\
        ShellfishOpenImages & 3 & 116 \\
        ThermalDogsAndPeople & 2 & 41 \\
        VehiclesOpenImages-416x416 & 5 & 200 \\
        \hline
      \end{tabular}
    }
  \vspace{-0.6em}
  \vspace{-0.6em}
  \label{tab:odinw13_stats}
\end{table}

\section{Additional Visualization Results}
\begin{figure*}[t]
  \centering
  \includegraphics[width=1.0\linewidth]{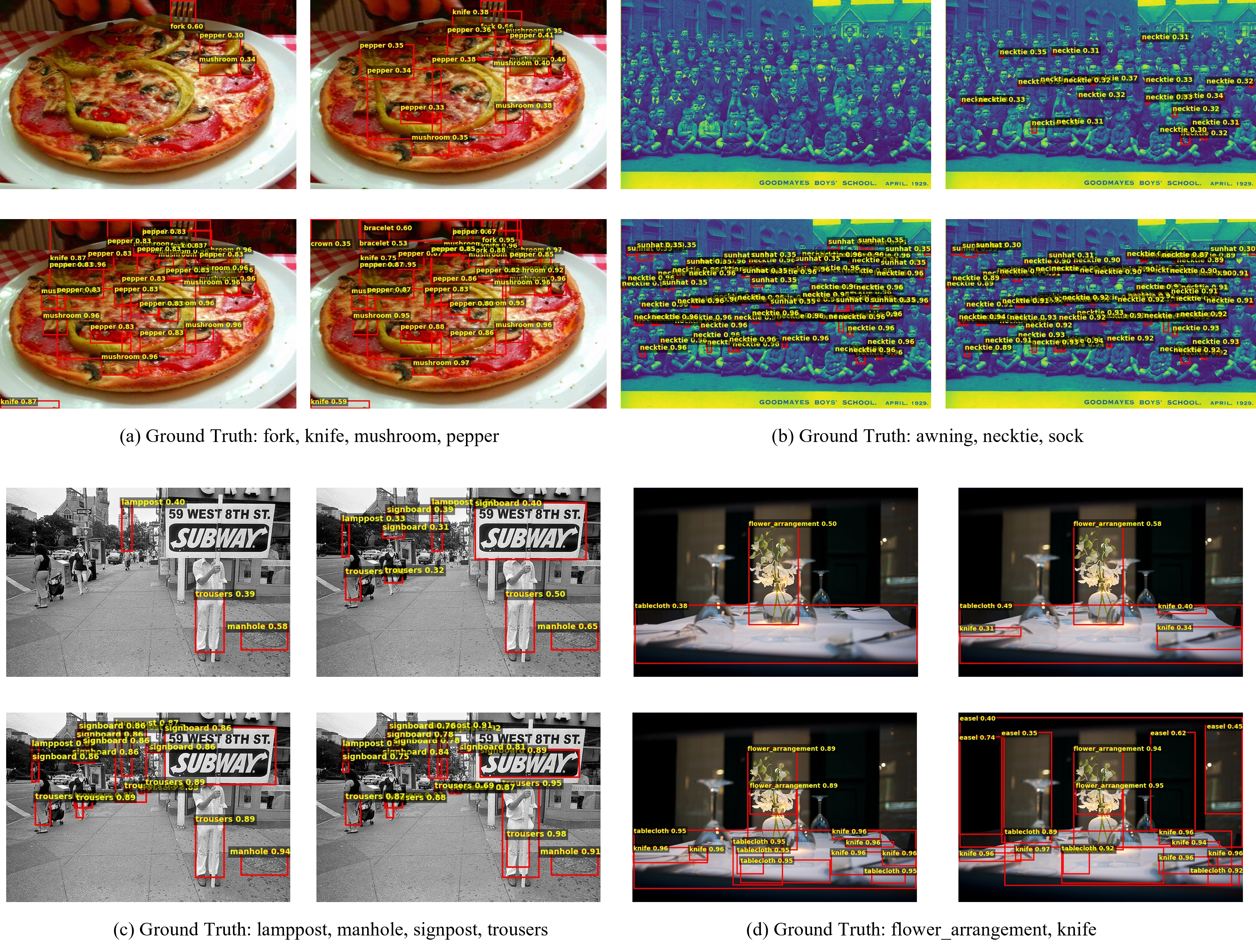}
  \caption{
Another qualitative comparison of detection results before and after applying DetRefiner.
Top: base detector (left) vs. base detector + DetRefiner (right).
Bottom: predictions based on the class vector (left) and patch vector (right).
DetRefiner suppresses overconfident false positives and recovers missed objects by combining global and local cues. For visualization, a box score threshold of 0.3 and an IoU threshold of 0.3 are applied for class-wise NMS on all images.
  }
  \vspace{-0.6em}
  \label{fig:qualitative_results2}
\end{figure*}

Figure~\ref{fig:qualitative_results2} illustrates how DetRefiner refines predictions from the base detector.
It suppresses overconfident false positives and boosts missed true positives using both global and local cues.
The bottom row indicates scene-level (class vector) and region-level (patch vector) calibration, which together improve open-vocabulary detection reliability.

\textbf{pizza scene} Fig.~\ref{fig:qualitative_results2}(a) shows a pizza image with several small toppings.
The base detector (top-left) assigns low confidence to many pepper and mushroom instances, whereas applying the class and patch vectors (bottom row) upweights boxes supported by global and local cues, so the full DetRefiner (top-right) yields consistently high confidence for most true toppings.

\textbf{Group photo}
Fig.~\ref{fig:qualitative_results2}(b) shows a crowded group photo with unusual color tones and many tiny neckties, socks, and awnings.
Although categories such as \textit{awning} and \textit{sock} remain missed because the original detector cannot detect them even with a zero-threshold setting, DetRefiner (top-right) reliably detects tiny neckties in the crowd.

\textbf{street scene} Fig.~\ref{fig:qualitative_results2}(c) presents a street scene with signboard, lamppost, trousers, and manhole.
Here DetRefiner uses the class vector to boost scene-consistent categories and the patch vector to further upweight boxes aligned with local structures, producing a more complete and reliable confidence distribution than the base detector.

\textbf{indoor table scene} Fig.~\ref{fig:qualitative_results2}(d) shows an indoor table with a flower arrangement, tablecloth, and knife.
While the base detector mainly scores the central flower highly, DetRefiner raises the confidence of table-related boxes whose features match knives and tablecloth regions, leading to dense and well-calibrated scores for the relevant objects.

\begin{figure*}[t]
  \centering
  \includegraphics[width=1.0\linewidth]{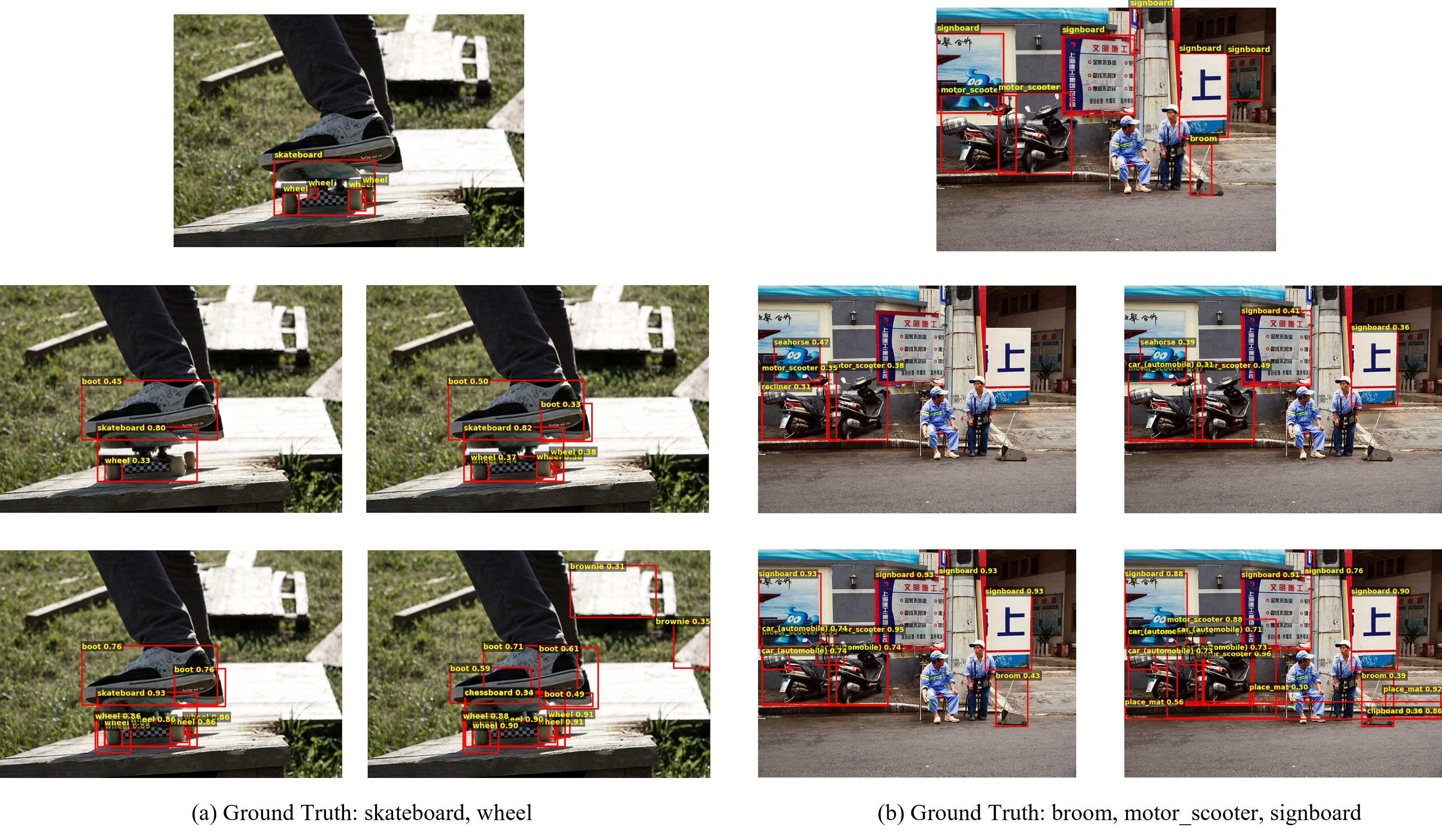}
  \caption{Additional qualitative success and failure cases. Top row: ground-truth bounding boxes. For each example, the middle row shows predictions from the base detector (left) and from the base detector with DetRefiner (right). The bottom row shows predictions from the class-vector branch (left) and patch-vector branch (right). For visualization, a box score threshold of 0.3 and an IoU threshold of 0.3 are applied for class-wise NMS on all images.}
  \vspace{-0.6em}
  \label{fig:qualitative_results3}
\end{figure*}

Figure \ref{fig:qualitative_results3} shows two additional scenes highlighting both the strengths and limitations of DetRefiner. For each example, the top row shows ground-truth boxes, the middle row compares the base detector (left) with DetRefiner (right), and the bottom row shows predictions from the class-vector branch (left) and the patch-vector branch (right).

\textbf{skateboard scene}
In Figure \ref{fig:qualitative_results3}(a), the base detector assigns low confidence to the wheels and tends to miss them.
DetRefiner successfully recovers the wheels with higher confidence, indicating that fused global–local cues can rescue missed objects.
At the same time, spurious boot predictions are actually reinforced and even newly introduced.
Because the objects detected as boot are visually ambiguous and can be interpreted as boots, both the class and patch branches assign relatively high probability to \emph{boot}, so the fused score becomes even larger than the base detector’s score and the false positives remain.

\textbf{street worker scene}
In Figure \ref{fig:qualitative_results3}(b), the base detector hallucinates a recliner and misses several signboards and a broom.
DetRefiner suppresses the recliner false positive and recovers some signboards, but still fails to detect the broom and some other signboards.
In addition, it introduces visually plausible yet incorrect predictions such as \textit{car\_automobile} around the scooter, where the local appearance is highly confounding.
Moreover, an overconfident \textit{seahorse} prediction is only slightly reduced and remains above the threshold.
These cases illustrate that DetRefiner is effective for moderate miscalibration and missed detections, but highly misaligned base scores can still dominate the final confidence and may even increase false positives for ambiguous classes.

\end{document}